\documentclass[pdflatex,sn-mathphys-num]{sn-jnl}

\usepackage{graphicx}%
\usepackage{multirow}%
\usepackage{amsmath,amssymb,amsfonts}%
\usepackage{amsthm}%
\usepackage{mathrsfs}%
\usepackage[title]{appendix}%
\usepackage{xcolor}%
\usepackage{textcomp}%
\usepackage{manyfoot}%
\usepackage{booktabs}%
\usepackage{algorithm}%
\usepackage{algorithmicx}%
\usepackage{algpseudocode}%
\usepackage{listings}%

\usepackage{tabularx}
\usepackage{makecell}
\usepackage{subcaption}
\usepackage{threeparttable}

\theoremstyle{thmstyleone}%
\theoremstyle{thmstyletwo}%

\theoremstyle{thmstylethree}%

\begin{document}

\title[Article Title]{Reducing Robotic Upper-Limb Assessment Time While Maintaining Precision: A Time Series Foundation Model Approach}


\author*[1]{\fnm{Faranak} \sur{Akbarifar}}\email{f.akbarifar@queensu.ca}

\author[1]{\fnm{Nooshin} \sur{Maghsoodi}}\email{nooshin.maghsoodi@queensu.ca}

\author[2]{\fnm{Sean P.} \sur{Dukelow}}\email{spdukelo@ucalgary.ca}

\author[3, 4]{\fnm{Stephen H.} \sur{Scott}}\email{steve.scott@queensu.ca}

\author[1]{\fnm{Parvin} \sur{Mousavi}}\email{mousavi@queensu.ca}

\affil*[1]{\orgdiv{School of Computing}, \orgname{Queen's University}, \orgaddress{\street{25 Union St.}, \city{Kingston}, \postcode{K7L 2N8}, \state{ON}, \country{Canada}}}

\affil[2]{\orgdiv{Hotchkiss Brain Institute}, \orgname{University of Calgary}, \orgaddress{\street{3330 Hospital Dr NW}, \city{Calgary}, \postcode{T2N 4N1}, \state{AB}, \country{Canada}}}

\affil[3]{\orgdiv{Centre for Neuroscience Studies}, \orgname{Queen's University}, \orgaddress{\street{18 Stuart St.}, \city{Kingston}, \postcode{K7L 3N6}, \state{ON}, \country{Canada}}}

\affil[4]{\orgname{Providence Care Hospital}, \orgaddress{\street{752 King St.}, \city{Kingston}, \postcode{K7L 4X3}, \state{ON}, \country{Canada}}}




\abstract{\textbf{Purpose:} Visually Guided Reaching (VGR) on the Kinarm robot yields sensitive kinematic biomarkers but requires 40–64 reaches, imposing time and fatigue burdens. We evaluate whether time series foundation models can replace unrecorded trials from an early subset of reaches while preserving reliability of standard Kinarm parameters.

\textbf{Methods:} We analyzed VGR speed signals from 461 stroke, and 599 control participants across 4- and 8-target reaching protocols. We withheld all but the first 8 or 16 reaching trials and used ARIMA, MOMENT, and Chronos models, fine-tuned on 70\% of subjects, to forecast synthetic trials. We recomputed four kinematic features of reaching (reaction time, movement time, posture speed, max speed) on combined recorded plus forecasted trials and compared to full-length references using ICC(2,1).

\textbf{Results:} Chronos forecasts restored ICC ($\geq 0.90$) for all parameters with only 8 real trials plus forecasts, matching the reliability of 24–28 recorded reaches ($\Delta ICC \leq 0.07$). MOMENT yielded intermediate gains, while ARIMA improvements were minimal. Across cohorts and protocols, synthetic trials replaced reaches without significantly compromising feature reliability.

\textbf{Conclusion:} Foundation-model forecasting can greatly shorten Kinarm VGR assessment time. For the most impaired stroke survivors, sessions drop from 4–5 minutes to about 1 minute while preserving kinematic precision. This forecast-augmented paradigm promises efficient robotic evaluations for assessing motor impairments following stroke.}




\maketitle

\section{Introduction}\label{sec:intro}

Upper-limb robotic devices are increasingly used in research laboratories to assess stroke-related impairments. These devices yield precise, quantitative performance measures that complement clinical scales \cite{Scott2011,balasubramanian2012robotic,Gassert2018,park2023robot}. An example is the \textit{Kinarm Exoskeleton Lab}, an interactive robotic device whose kinematic measures correlate with traditional clinical scales such as the Fugl–Meyer Assessment~\cite{FuglMeyer1975} and the Chedoke–McMaster Stroke Assessment (CMSA)~\cite{Gowland1993} \cite{SCOTT1999119,Simmatis2020,akbarifar2022computer}. Kinarm Standard Tests (KST) ~\cite{Simmatis2017, kinarm} provides a suite of behavioural tasks to assess a broad range of sensory, motor and cognitive functions and has been used to assess impairments associated with stroke \cite{Coderre2010, pmid24693877, Bourke2016ARB, mochizuki2019movement} and a broad range of other neurological diseases and injuries \cite{mang2019robotic,simmatis2020epilepsy, vanderlinden2022acute, bray2024robotic}. During each task, the robotic system records precise upper limb location and velocity data, quantifies task-specific spatiotemporal features, combines them into a single Task Score, and leverages statistical models to compare a participant's performance against healthy individuals, considering factors such as age, sex, and handedness \cite{SCOTT1999119, Scott2011, Simmatis2020, scott2022assessment}. 

Visually Guided Reaching (VGR) task \cite{Coderre2010} is commonly used to quantify upper limb motor function through spatiotemporal features that are categorized into 5 attributes of sensorimotor control:  upper-limb postural control, reaction time, initial movement, movement corrections and total movement metrics, and are sensitive to identify mild impairment \cite{akbarifar2022computer}. A practical limitation is duration: a full VGR block presently requires 40–64 reaches ($\approx$ 4–5 minutes), which constrains throughput and increases fatigue. Shorter sessions reduce subject fatigue and overall assessment time, improve throughput and scheduling, and can make robotic assessment more tolerable for individuals with severe motor and cognitive impairments.
In the current paper, we ask whether a forecast-augmented VGR session can maintain measurement precision and clinical decisions while substantially reducing on-robot time.

Time reduction on Kinarm has been pursued at three complementary levels. (i) Task redundancy: Quantitative similarity between Object Hit (OH)~\cite{Bourke2016ARB} and Object Hit and Avoid (OHA)~\cite{pmid24693877} has been tested by predicting the OH parameter set from OHA with Fast Orthogonal Search (FOS)~\cite{korenberg1988identifying} and then classifying stroke vs. control using the predicted OH parameters; classification performance was comparable to using the measured OH parameters, supporting removal of one member of the OH/OHA pair in batteries where both appear \cite{mostafavi2015evaluation}. (ii) Protocol flow: Using non-linear hierarchical ordering~\cite{bart1973ordering} on discretized (pass/fail) task outcomes, an inferred partial order lets early tasks predict downstream outcomes with high confidence, enabling conditional skipping of later tasks. In one cohort this yielded 97\% confidence for a subgroup and $\geq$8 minutes saved \cite{mostafavi2014hierarchical}; generalized across five standard tasks, a hierarchical task-selection strategy that combines ordering, parameter ranking, and cross-task prediction reported large ($\geq$50\%, up to 90\%) time reductions \cite{Mostafavi2017}. (iii) Within-task VGR: Reduced-trial VGR protocols (fewer reaches) were evaluated by reestimating the VGR parameters under each scheme and validating diagnostic utility with an SVM classifier; sensitivity, specificity, and accuracy were nearly identical to the full protocol, indicating substantial recording-time savings at minimal loss in classification performance \cite{mostafavi-vgr-reduction}.

 The strategies above shorten assessment time by removing tasks, reordering tasks to skip later ones, or recording fewer trials. Here, we take a complementary route: retain the existing analytics and scoring pipeline, but replace a subset of recorded trials with model-generated forecasts conditioned on the subject’s early trials. This keeps the protocol and outcome measures intact while cutting the recording time and preserving downstream interpretability (the same speed-derived parameters are computed on recorded+forecasted trials). We adopt the state-of-the-art time series foundation models because they learn transferable priors on extensive time series data and are adapted with limited domain data \cite{bommasani2021opportunities}. In our experiments we instantiate this idea with two representative forecasters, Chronos \cite{Ansari2024Chronos} and MOMENT \cite{Goswami2024MOMENT}, compared with a classical baseline (ARIMA) \cite{box2015time}. To our knowledge, these models have not been evaluated for robotic stroke assessment nor for forecasting reaching patterns in VGR. We therefore pose the unrecorded trials as conditional sequence generation given early trials as context and measure success by agreement of downstream KST parameters computed on real + forecasted trials with those from full recordings.

We introduce a forecast-augmented version of the Kinarm VGR assessment that preserves the standard task design and analytics while reducing recorded trials; formulate a time-series, direction-conditioned forecasting approach that generates subject-specific trials from a short recorded context without altering canonical VGR features or their computation \cite{kinarm,Coderre2010}; design a subject-level, multi-site evaluation across stroke and control cohorts that compares forecast-augmented sessions to full-length references using established reliability metrics (e.g., ICC); and quantify uncertainty by combining variability from subject sampling with variability due to repeated selection of forecasted trials. The remainder of this article describes the dataset and preprocessing, then details the forecasting setup and evaluation protocol; we next present results and ablations, and conclude with clinical implications, limitations, and future work.

\section{Methods}\label{sec:methods}

\subsection{Participants}
Participants with stroke and healthy volunteers were recruited from the communities of Kingston, Ontario and Calgary, Alberta. Participants with stroke were recruited from the inpatient acute stroke or stroke rehabilitation units at the Foothills Medical Centre, the Dr. Vernon Fanning Care Centre in Calgary, Alberta, Canada, and Providence Care, and St Mary’s of the Lake Hospital, Kingston, Ontario, Canada. Participants with stroke were assessed using a variety of clinical measures, including the Chedoke-McMaster Stroke Assessment~\cite{Gowland1993}. This study was reviewed and approved by the University of Calgary Conjoint Health Research Ethics Board and the Queen’s University Research Ethics Board. All participants gave their written informed consent to have their data collected for research purposes before performing the assessment. The inclusion criteria for participants with stroke included a first-identified stroke within the last 35 days and being at least 18 years old. Individuals were excluded if they had other neurological disorders (e.g. Parkinson’s, Multiple Sclerosis), upper limb orthopedic impairments, difficulties understanding tasks instructions or signs of apraxia \cite{ vanbellingen2011new}. Healthy control participants fulfilled the same inclusion and exclusion criteria, except had no stroke history, and did not have any known neurological or upper arm musculoskeletal  conditions. 

\subsection{Apparatus}
All data in this study were gathered using the Kinarm Exoskeleton Lab \cite{kinarm}\cite{SCOTT1999119}, an interactive robotic system equipped with an integrated virtual reality system designed to measure upper limb sensorimotor function. The participants were positioned in an adjustable-height chair and the robotic linkages were adjusted to align the robot's joints with the participants' shoulder and elbow joints, enabling them to both flex and extend their shoulders and elbows while supporting the arm against gravity (Figure \ref{fig:1a}). The virtual reality display in this interactive robotic system provided visual feedback aligned with the horizontal workspace with a physical barrier obscuring direct view of the participants' arms.

\subsection{VGR task}

\begin{figure*}[!htbp]
  \centering
  \subfloat[Kinarm exoskeleton robot]{%
    \includegraphics[width=0.5\textwidth]{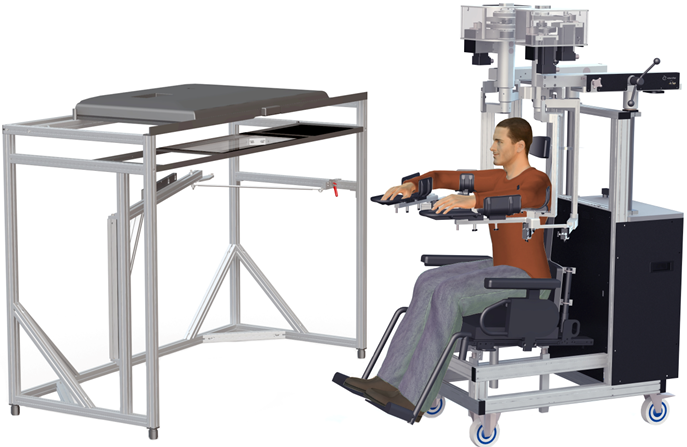}\label{fig:1a}}%
    \hfill
  \subfloat[VGR virtual workspace]{%
    \includegraphics[width=0.4\textwidth]{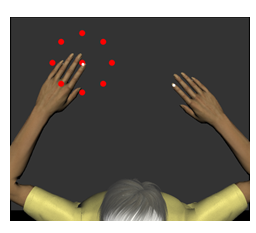}\label{fig:1b}}\\[1ex]
  
  \subfloat[A sample speed trial]{%
    \includegraphics[width=0.9\textwidth]{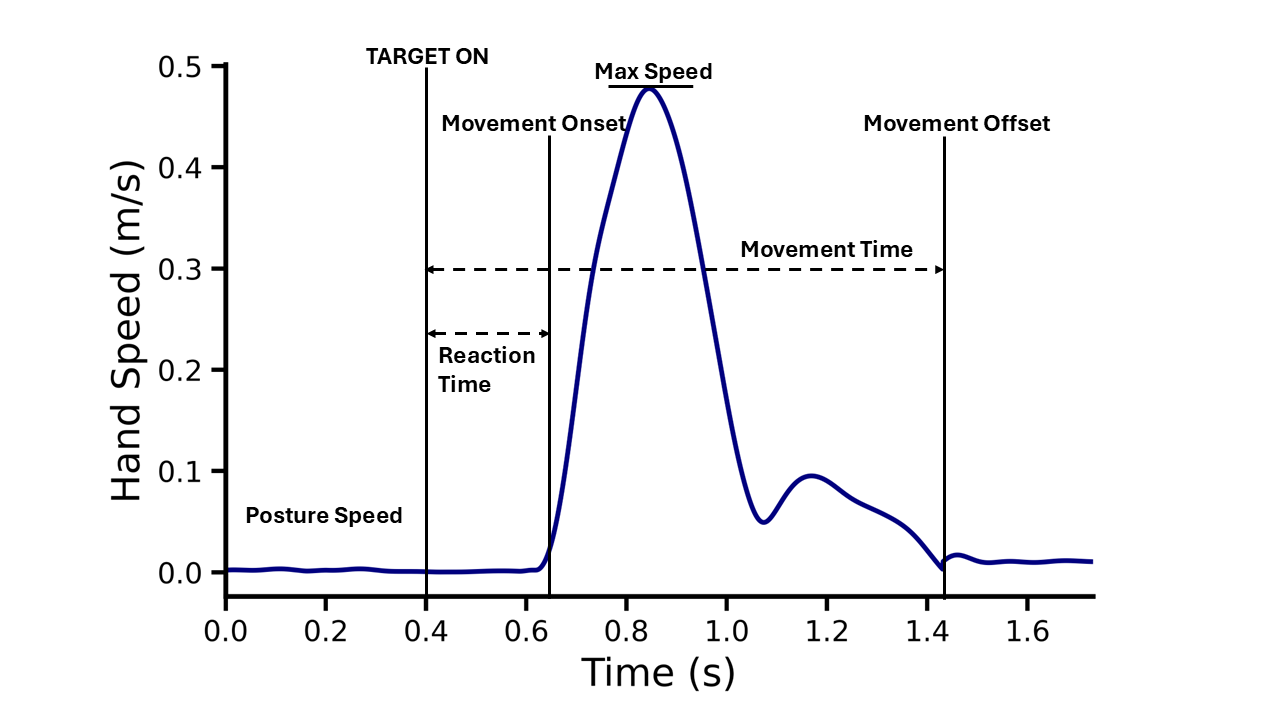}\label{fig:1c}}\\[1ex]
  \subfloat[Forecast-augmented assessment pipeline]{%
    \includegraphics[width=0.8\textwidth]{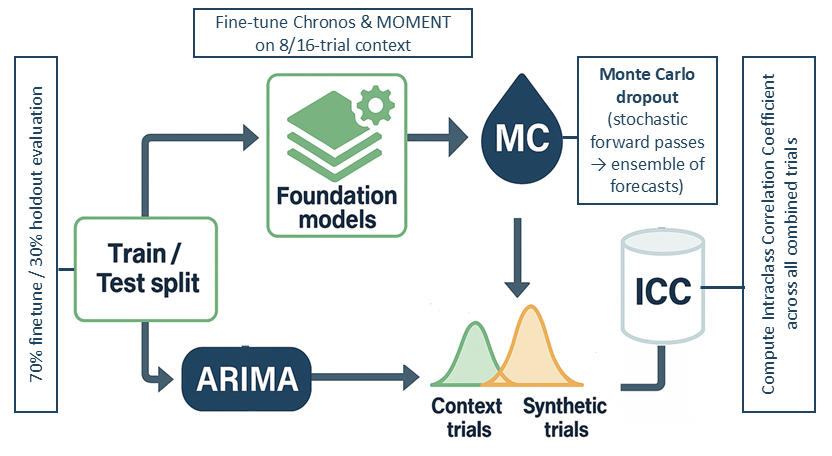}\label{fig:1d}}

  \caption{Apparatus and task.  
    (a) Participant in the Kinarm Exoskeleton Lab.  
    (b) Virtual workspace for the
    Visually Guided Reaching (VGR) task: the hand cursor moves from the
    central start target to one of the eight peripheral targets that
    appear in random order.  
    (c) Representative control-trial hand-speed trace illustrating computation of the four selected Kinarm parameters—posture speed, reaction time (TARGET ON→Movement Onset), movement time (Movement Onset→Movement Offset), and max speed— with TARGET ON, Movement Onset and Movement Offset marked.
    (d) Study pipeline: an 8‑trial
    context is fed into a fine‑tuned foundation model (Chronos or
    MOMENT) with Monte‑Carlo dropout to generate additional synthetic
    trials; real + forecasted trials are then used to recompute
    KST‑defined speed parameters and their intraclass correlation
    coefficients (ICC).}
  \label{fig:figure1}
\end{figure*}

In the VGR task, the integrated virtual reality system provides visual feedback on hand position and spatial objects aligned within the horizontal workspace (Figure \ref{fig:1b}).
Over the years, several closely related variants of the Kinarm VGR task have been developed. In each version, participants view their hand as a bright circle on the exoskeleton’s co-located virtual‐reality display and reach toward one of several 10 cm‐distant peripheral targets (Figure \ref{fig:1b}). In the earliest ``reachout\_std" 8-target protocol (P1), eight equiangular targets (45° spacing) appear once per block over eight blocks (64 trials), with two ``catch" trials per block (no target) to assess postural stability; only the outward reaches are analysed \cite{Coderre2010}. A second 8-target variant (P2: ``8-target 10 cm reach") preserves the same spatial layout and total trial count but implements updated hold-times and inter-trial intervals in the Kinarm software. More recently, the ``4-target In\&Out 10 cm" protocol (P3) reduced the workspace to four targets at 0°, 90°, 180° and 270°: here each peripheral direction is reached five times in a pseudo-random order, followed by a return to the central target (when the central target re-illuminates), and four catch trials (no target) are interspersed, yielding up to 40 outbound–inbound movements per session. Across all variants, hand speed and position are sampled at 1 kHz. Fourteen spatiotemporal features are extracted per trial (see \cite{Coderre2010} for full definitions) and form the basis for normative Task Scores. In the present study, we evaluate forecast quality using four task parameters (posture speed, reaction time, movement time, and max speed) within our reliability and forecasting analyses.

\begin{table}[t]
  \centering
  \small
  \caption{Number of control and participants with stroke by VGR task variant.}
  \label{tab:variant_counts}
  \begin{tabular}{lccc}
    \toprule
    \textbf{Task variant} & \textbf{Control (n)} & \textbf{Stroke (n)} & \textbf{Max trials per participant}\\
    \midrule
    P1: reachout\_std (8-target)      & 93 & 21 & 64\\
    P2: 8-target 10 cm reach          & 45 & 337 & 64\\
    P3: 4-target In\&Out 10 cm RT     & 461 &  73 & 40\\
    
    \bottomrule
  \end{tabular}
\end{table}

\subsection{Dataset formation}
\label{subsec:data-prep}

For every participant we extracted the hand speed signal of the more affected limb for participants with stroke and the non-dominant limb in controls—from the raw Kinarm sensor files (1 KHz sampling).

Data were collected over several years under the VGR
protocols mentioned in Table \ref{tab:variant_counts}. To obtain a temporally
consistent snippet from every file we anchored each trial to the
\textsc{TARGET\_ON} event, extracted a window beginning
200 ms \emph{before} that event and ending at
\[
t_{\text{stop}}
  =  \textsc{TARGET\_ON}
     + \text{ReactionTime}
     + \text{TotalMovementTime},
\]
where ReactionTime and TotalMovementTime are provided
per trial in the Kinarm log.  This window spans late movement preparation,
and movement execution for all participants regardless of
protocol year. For participants with stroke that could not reach the end target, the window ended at the end of trial.

Each extracted speed trace was linearly resampled to a fixed
$L = 64$ points—first, to normalise trial length across subjects whose
movements naturally vary in duration, and second, because the Chronos
family of time‑series foundation models shows best performance when
forecast windows have lengths of up to 64 samples
\cite{Ansari2024Chronos}.

Following the KST manual \cite{kinarm}, we compute four Kinarm parameters per trial and use them in all downstream reliability analyses. Panel (c) of Figure~\ref{fig:figure1} annotates a representative speed signal with the event times used for each definition: posture speed — the median hand speed during the stationary hold at the start; reaction time — the interval from \textsc{TARGET\_ON} to \textsc{Movement Onset}; movement time — the interval from \textsc{Movement Onset} to \textsc{Movement Offset}; and maximum speed — the peak hand speed between \textsc{Movement Onset} and \textsc{Movement Offset}. For a given set of trials, each parameter is summarized by the mean or median of its per-trial values, depending on the parameter.

\subsection{Forecasting Models}

We compare a statistical baseline (ARIMA) with two foundation model  forecasters (MOMENT, Chronos) for conditional generation of unrecorded reaches. Each model consumes a context of \(c\in\{8,16\}\) recorded reaches, where every reach is a hand speed sequence resampled to \(L=64\) samples (Section~\ref{subsec:data-prep}), and outputs a 64-sample forecast.

\paragraph{ARIMA}
We use a classical ARIMA\((p,d,q)\)~\cite{box2015time} baseline on each subject’s concatenated context signal \(y_t\) (\(t=1,\ldots,512\), i.e., 8 trials \(\times\) 64 samples). For each subject we choose \((p,d,q)\) by minimizing AICc over a small grid \(p,q\in\{0,1,2,3\}\) and \(d\in\{0,1\}\); if models tie, we prefer the lower order. After fitting the selected model, we generate multiple stochastic forecast paths by simulating future innovations \(\varepsilon_t \sim \mathcal{N}(0,\hat{\sigma}^2)\) and propagating them through the fitted ARIMA state. Concretely, for a requested \(K\) synthetic trials (e.g., \(K\in\{8,16,24,\ldots\}\)), we forecast a horizon \(h=64K\) samples and draw \(M\) independent paths \(\{\hat{y}_{512+1:512+h}^{(m)}\}_{m=1}^{M}\). Each length-\(h\) path is then partitioned into \(K\) non-overlapping 64-sample segments in chronological order to serve as synthetic trials, which we append to the 8-trial context before recomputing the four VGR parameters. Unlike MOMENT/Chronos, the ARIMA baseline is direction-agnostic (no conditioning on target direction); its role is to provide a transparent statistical comparator with uncertainty coming from the Gaussian innovations \(\varepsilon_t \sim \mathcal{N}(0,\hat{\sigma}^2)\).

\paragraph{MOMENT}
Our goal was to forecast a subject’s next reach \emph{for a specified direction} using that same subject’s prior reaches. We adapt \texttt{MOMENT-1-small} for single-trial, direction-conditioned forecasting \cite{Goswami2024MOMENT}. Each subject contributes 9 hand speed trials (1 channel, 64 samples): 8 as context (one per direction) and 1 representative as the forecast target; directions are encoded as $d\in\{0,\ldots,7\}$. A subject-level sampler yields batches of one subject so each optimization step processes  $\text{context}\in\mathbb{R}^{1\times 8\times 64}, \text{context\_dirs}\in\mathbb{Z}^{8}, \text{forecast}\in\mathbb{R}^{1\times 64}, and \text{forecast\_dir}\in\mathbb{Z}$. We freeze the backbone embedder and also train a lightweight aggregator head (dropout p{=}0.1) that: (i) encodes each context trial; (ii) applies a normalized mask that keeps only embeddings whose direction matches \text{forecast\_dir}; (iii) concatenates the masked average with a learned embedding of \text{forecast\_dir}; and (iv) projects to a 64-sample forecast. We minimize DTW loss with Adam (lr $5{\times}10^{-4}$, weight decay $10^{-5}$), early stopping (patience 15), and a max of 100 epochs. Data splits are at the subject level to prevent leakage. At test time, dropout remains active in the head to draw 8 Monte Carlo \cite{gal2016dropout} samples per direction.

\paragraph{Chronos}

We fine-tuned a Chronos forecaster \cite{Ansari2024Chronos} (\texttt{chronos-t5-small}) on resampled hand-speed sequences using an 
8×64 context and a 64-token future target. To condition forecasts on movement direction \emph{without} altering Chronos’s numeric (uniform-bin) tokenizer, we reserved eight additional special tokens in the vocabulary and used them as direction controls, \texttt{[DIR\_0]}…\texttt{[DIR\_7]}. For each training instance we prepended the matching \texttt{[DIR\_d]} to the encoder input, and used the next 64 tokens as labels (teacher forcing) with a deterministic last-window split per series. Training used HuggingFace \texttt{Trainer} with AdamW (fused), learning rate $1 \times 10^{-4}$
, per-device batch size 32 with gradient accumulation 2, logging every 500 steps, evaluation/checkpointing every 2,000 steps (keeping the latest 3), and early stopping (patience=5). At inference, we prepend the requested direction token and draw 64 stochastic sample paths via top-k decoding; decoded forecasts (de-quantized using the learned scale) are then combined with the eight recorded context trials to compute KST features and ICC.

\subsection{Forecast Experiments}\label{subsec:experiments}

In the context of time series forecasting, a ``context" refers to a known initial portion of sequential data provided to the model, serving as historical information from which future points are predicted. For our study, the context specifically consisted of a subset of early recorded trials selected to represent every unique movement direction once (8 reaches), providing a representative initial snapshot of each participant's reaching behavior. This initial set of context trials enables the forecasting model to learn subject-specific temporal patterns and general movement dynamics, thus allowing reliable prediction of subsequent, unrecorded reaches.

To ensure the context contained at least one exemplar of each movement direction, we systematically selected the first eight trials with unique direction codes in the following way. For the 8-target protocols (P1 and P2), we selected the first occurrence of each of the eight unique target directions, yielding eight context trials. In the 4-target protocol (P3), we similarly selected the first ``reach‐out" trial and “return” trial for each of the four targets, again totalling eight context trials. We did a similar process for selecting 16 trials based on movements associated with the first two presentations of each target.

Trials beyond these context trials were treated as future events for forecasting. Participants with fewer than eight valid trials were excluded from the primary (8-trial) analysis. For the 16-trial sensitivity analysis, we analogously required at least 16 valid trials.

After preprocessing, the dataset is represented by

\[
X^{(i)} \in \mathbb{R}^{8 \times 64},
\]

where the rows are context trials ordered by target direction and the
columns are resampled speed samples.  The accompanying labels are: stroke/control status, CMSA score, protocol identifier (P1–P3), trial‑level metadata (reaction time, movement time, etc.).

We merged our stroke and control datasets and conducted a randomized 70–30 train-test split, ensuring that each participant's data appeared exclusively in one set. The foundation models, MOMENT and Chronos, were fine-tuned on the training set, with each trial uniformly resampled to a fixed length of 64 samples for compatibility with model input constraints. Forecasting began from eight context trials representing the first trials performed in each direction (two trials per direction for the 4-target variant, one trial per direction for the 8-target variants).
We created multiple forecast groups using \textbf{ARIMA} (statistical baseline) \cite{box2015time}, \textbf{MOMENT} (transformer‐based foundation model) \cite{Goswami2024MOMENT}, \textbf{Chronos} (tokenized foundation model) \cite{Ansari2024Chronos}.
Reliability of kinematic features from combined real+forecasted trials for each method was assessed using the ICC, as detailed in the following subsection.

To assess robustness to the amount of historical context, we repeated all preprocessing, modeling, and evaluation steps with a 16-trial context, i.e., \(X^{(i)} \in \mathbb{R}^{16 \times 64}\). All other settings were unchanged. Comparative results for 8- and 16-trial contexts are reported in the Results.

\subsection{Evaluation metrics}\label{subsec:eval}
As stated in subsection \ref{subsec:data-prep}, we compute per-trial values for four KST speed-based parameters. For each parameter and cohort, we summarize trial-level repeatability at the subject level using the ICC, and we report the mean and 95\% confidence interval across subjects.

We estimate agreement between paired measurements using a two-way random-effects, absolute-agreement intraclass correlation, ICC(2,1) \cite{ShroutFleiss1979,McGrawWong1996}. 
For interpretability we follow Koo and Li (2016): ICC $<$ 0.50 = poor, 0.50–0.75 = moderate, 0.75–0.90 = good, and $>$ 0.90 = excellent \cite{KooLi2016}.

Error bars in our figures reflect two sources of variability, depending on the condition.
For the recorded-only baseline, we estimate uncertainty via a bootstrap over subjects (with replacement, $B{=}1000$) and report the 2.5--97.5th percentiles of ICC across bootstrap replicates.
For forecast-augmented points, we quantify forecast selection variability by repeatedly sampling the required number of forecasted trials from each subject's dropout-generated forecast pool and recomputing ICC; we display the mean across repeats with $\pm$ one standard deviation as error bars.

To assess the effect of adding forecasted trials, we first construct a recorded-only benchmark curve.
For each trial count $X$, we compute a subject-level metric (e.g., the mean reaction time over $X$ recorded trials) and then the ICC between the $X$-trial metric and the complete-trial metric; uncertainty is obtained by bootstrapping subjects ($B{=}1000$), as described above.

For forecast-augmented protocols, we fix a context size (e.g., $c{=}8$ recorded trials) and add $k\in\{0,8,16,\ldots\}$ forecasted trials per subject.
Forecasts are produced by keeping dropout active at inference to form a forecast pool per subject; for each $k$ we draw $R$ independent selections from this pool, recompute the metric on the $(c{+}k)$ trials, and evaluate ICC against the complete-trial metric.
Plotted squares show the mean ICC across the $R$ selections with vertical error bars indicating the standard deviation (forecast-selection variability).
Improvements are summarized by the absolute change $\Delta\mathrm{ICC}$ and the percent change relative to the recorded-only baseline, both averaged across subjects.
For model benchmarking, we repeat the identical procedure for each forecaster (MOMENT, Chronos) and the ARIMA baseline using the same preprocessing and splits.

\section{Results}\label{sec:results}

\subsection{Participants} \label{sub:partcipants}
The dataset comprised robotic assessments from 461 participants with stroke and 599 neurologically healthy controls.  Key demographic and clinical characteristics for both groups appear in Table \ref{tab:tab_demo}.  Right-handedness predominated (90\% in controls, 92\% in participants with stroke), with mixed handedness recorded in only five individuals (four controls, one participant with stroke) according to the Edinburgh Handedness Inventory \cite{veale2014edinburgh}.  Based on the Chedoke–McMaster Stroke Assessment, 73\% of participants with stroke showed motor deficits in their more-affected limb.  Every participant completed one of three VGR task variants, but the distribution was uneven: control participants were mainly tested on the 4-target version, whereas participants with stroke mainly undertook the 8-target protocols (Table \ref{tab:variant_counts}).  Consequently, all analyses are reported separately by \textit{cohort} (control vs.\ stroke) and by \textit{protocol} (4- vs.\ 8-target).

\begin{figure}[!htbp]
  \centering
  
  \newlength{\toprowheight}
  \setlength{\toprowheight}{0.3\textheight} 

  \begin{subfigure}[t]{0.48\linewidth}
    \centering
    \includegraphics[height=\toprowheight]{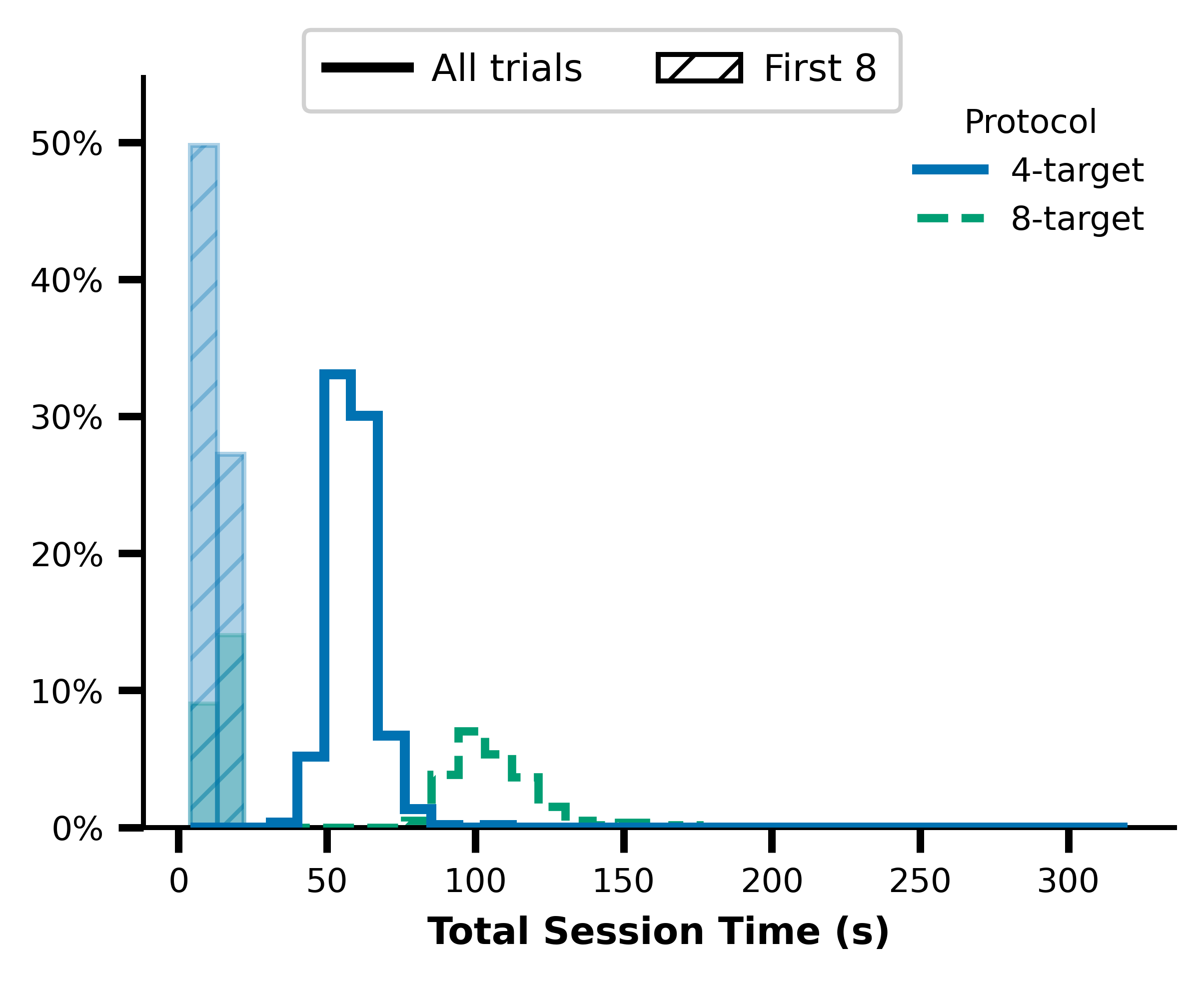}
    \subcaption{Control — total session times (histogram)}
    \label{fig:time-protocol-control}
  \end{subfigure}\hfill
  \begin{subfigure}[t]{0.48\linewidth}
    \centering
    \includegraphics[height=\toprowheight]{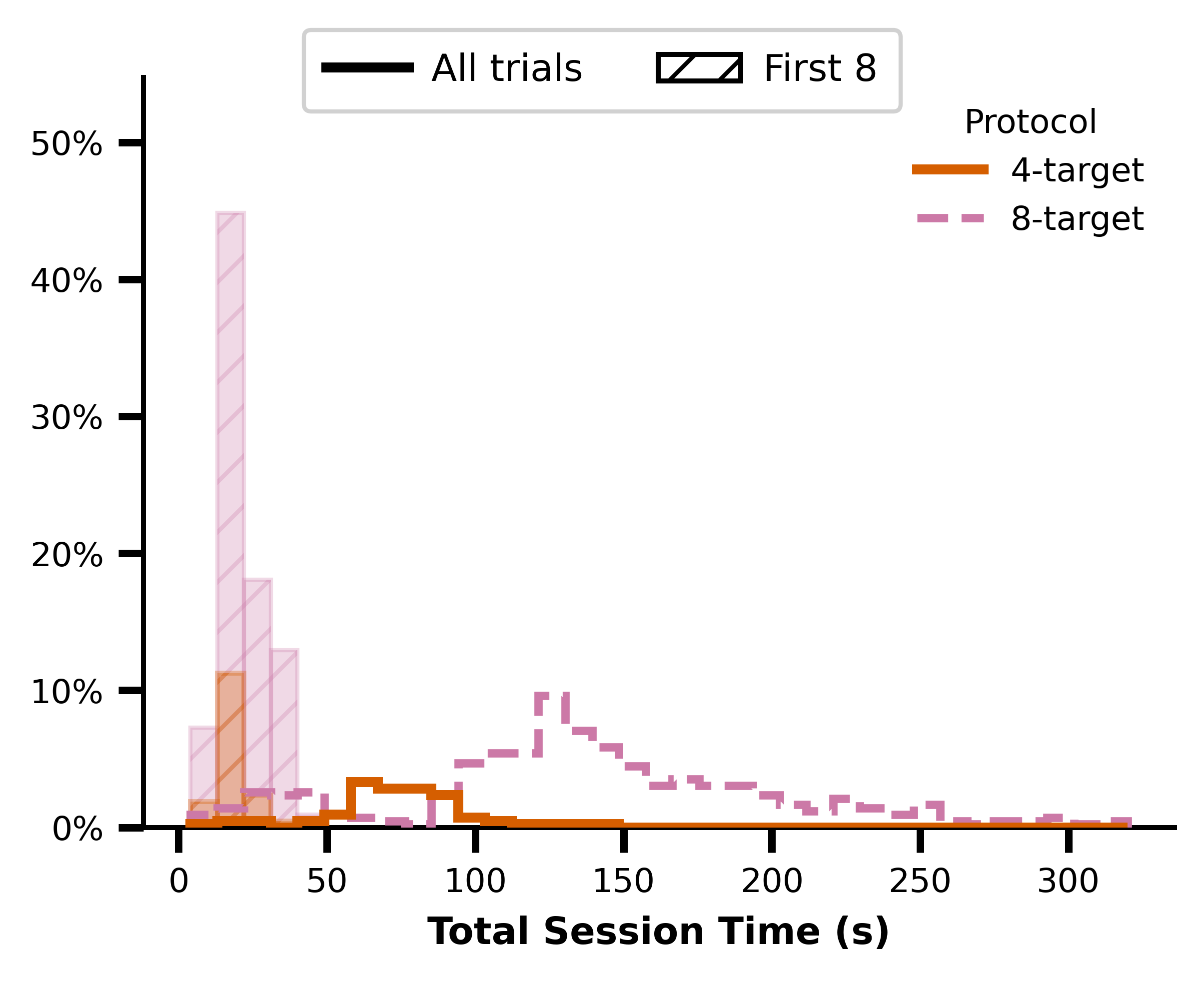}
    \subcaption{Stroke — total session times (histogram)}
    \label{fig:time-protocol-stroke}
  \end{subfigure}

  \vspace{0.4em}

  \begin{subfigure}[t]{0.7\linewidth}
    \centering
    \includegraphics[width=\linewidth]{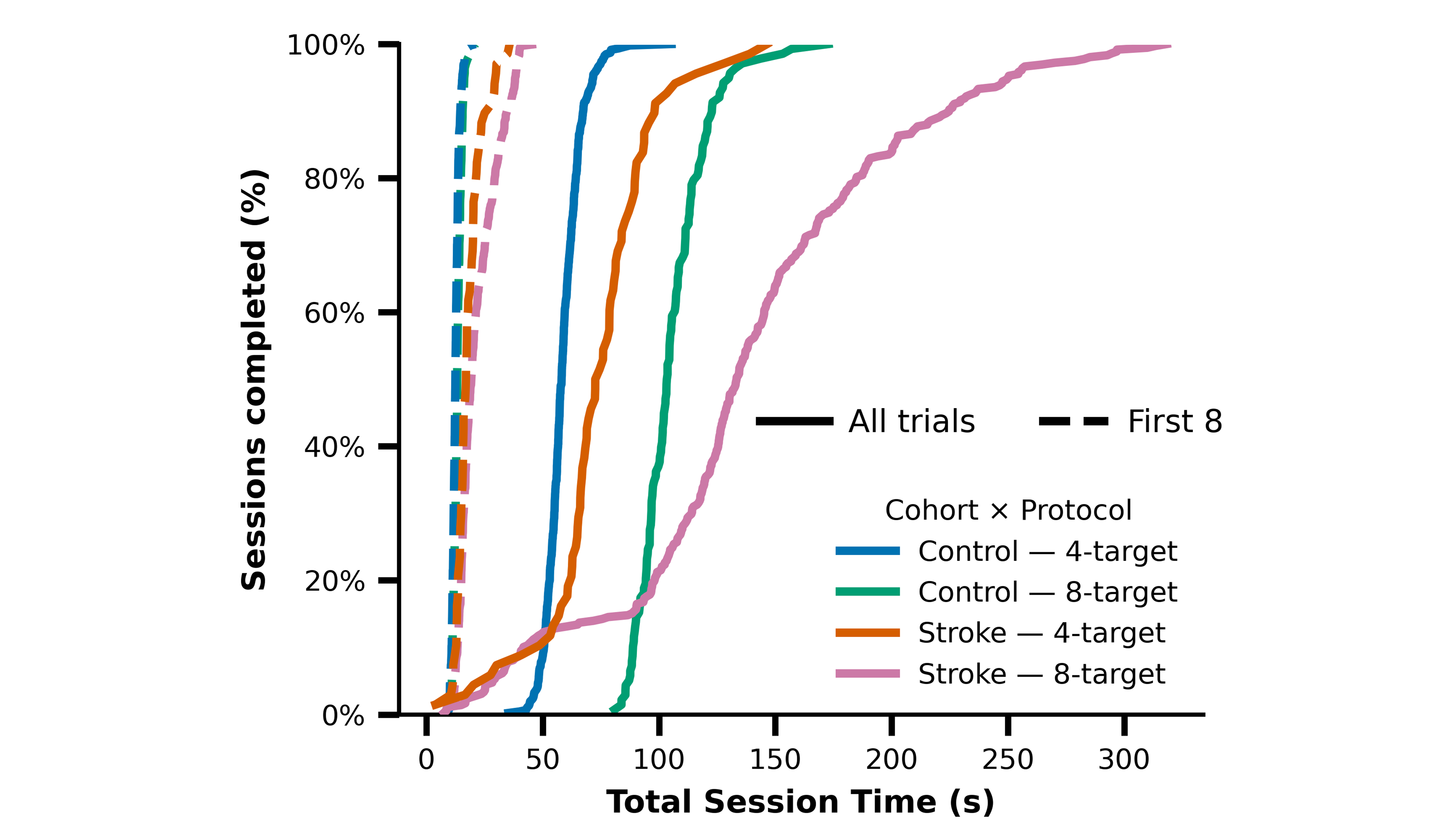}
    \subcaption{ECDF of total session time by cohort and protocol}
    \label{fig:time-protocol-ecdf}
  \end{subfigure}

  \caption{Session duration with all trials vs.\ first eight, by cohort and protocol. 
    (a) Control and (b) Stroke: normalized histograms of total VGR session time (sum of trial durations), overlaid by protocol (4-target vs.\ 8-target). Within each panel, solid outlines show all recorded trials and hatched bars show first eight trials only; binning is shared and percentages sum to 100\% per cohort. 
    (c) Empirical cumulative distribution (ECDF) of total session time by cohort\,$\times$\,protocol. Solid curves use all trials; dashed curves use the first eight. 
    Across cohorts and protocols, the first-8 condition shifts distributions leftward and steepens the ECDF, indicating markedly shorter sessions, while preserving the expected ordering (8-target $>$ 4-target; stroke $>$ control).}
    \label{fig:time-comp}

\end{figure}

To quantify the time cost of full-length recordings, we computed each participant’s total session time as the sum of trial durations, and, for comparison, a first-8 total by summing only the first eight recorded trials. Figure~\ref{fig:time-comp} summarizes these distributions by cohort and protocol. Panels (a–b) show that trimming to eight trials shifts the session-time histograms markedly leftward and narrows their spread for both cohorts, whereas the stroke distributions remain right-shifted relative to controls (\textit{i.e.}, group differences are preserved rather than collapsed by truncation. Panel (c) (ECDFs) highlights the practical gain: with the first-8 design, all sessions finish within about a minute across protocols, whereas the all-trials curves extend well beyond two to five minutes, especially for participants with stroke in the 8-target task.

\begin{table}[!htbp]
\centering
\caption{Demographic and clinical data for control and participants with stroke.}
\label{tab:tab_demo}
\small
\begin{tabularx}{\textwidth}{Xcc}
  \toprule
                      & \textbf{Control (n=599)} & \textbf{Stroke (n=461)} \\
  \midrule
  Age                 & 46 (18–93)               & 62 (18–92)               \\
  \addlinespace
  \addlinespace
  Sex                 & 167 M, 201 F             & 221 M, 115 F, 1 O        \\
  \addlinespace
  \addlinespace
  Dominant hand       & 330 R, 34 L, 4 A         & 310 R, 26 L, 1 A         \\
  \addlinespace
  \addlinespace
  Days since stroke   & —                        & 16 (1–84)                \\
  \addlinespace
  \addlinespace
  Type of stroke      & —                        & 290 I, 45 H, 2 U         \\
  \addlinespace
  \addlinespace
  CMSA [1–7]          &                          &                          \\
  \addlinespace
    \hspace{10pt}More affected arm        & —               & [16, 40, 48, 23, 58, 56, 96] \\
  \addlinespace
    \hspace{10pt}Less affected arm      & —               & [0, 0, 0, 1, 12, 61, 263]    \\
  \bottomrule
\end{tabularx}

\vspace{1ex}
{\footnotesize
\textbf{Notes.} Data are mean (range).  
CMSA = Chedoke–McMaster Stroke Assessment;  
M = male; F = female; O = other;  
R = right; L = left; A = ambidextrous;  
I = ischemic stroke; H = hemorrhagic stroke; U = unknown.  

}
\end{table}

\begin{figure*}[!htbp]
  \centering
  \begin{subfigure}[b]{0.48\linewidth}
    \centering
    \includegraphics[width=\linewidth]{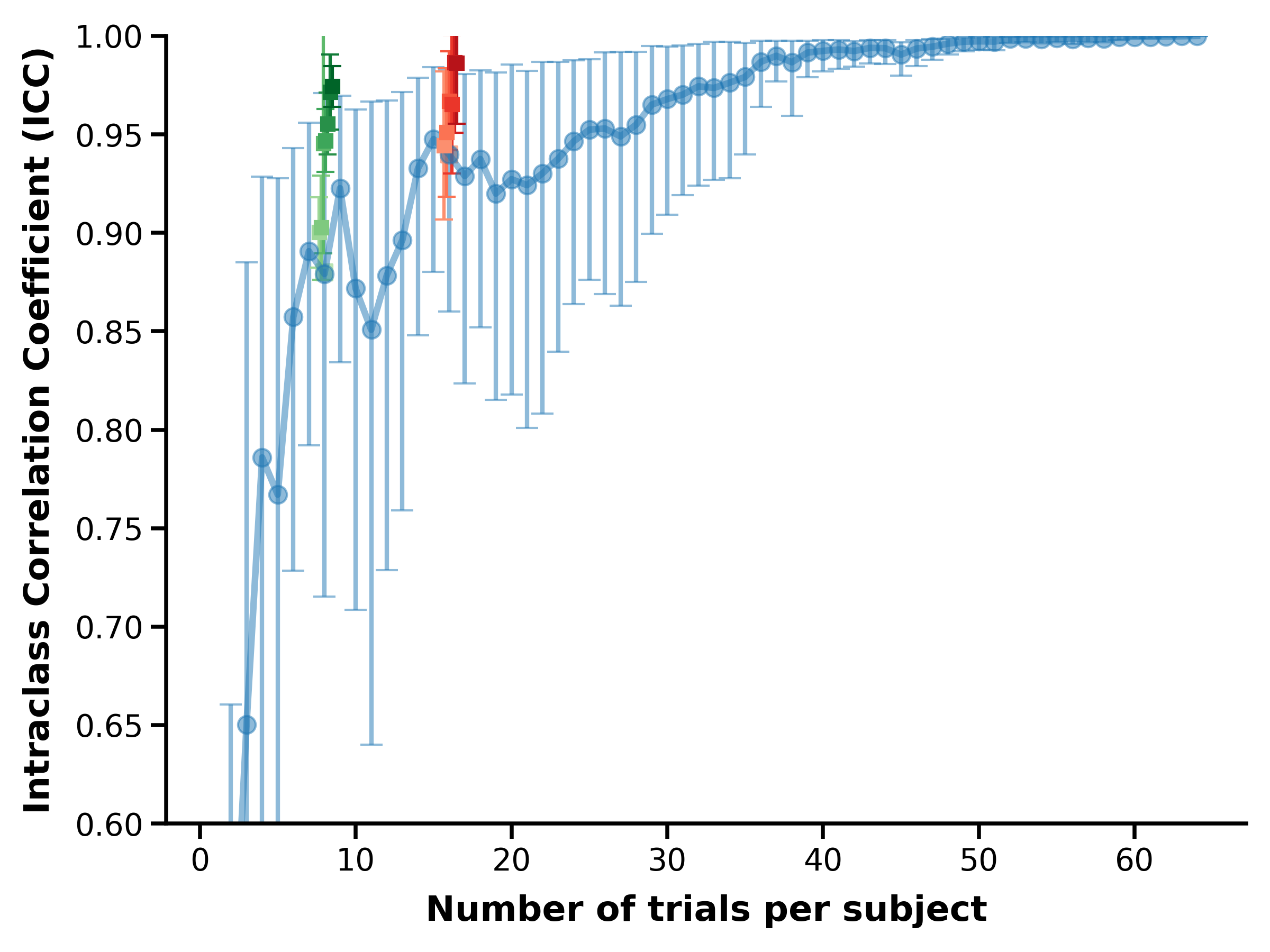}
    \caption{Control, 8-target protocol}
    \label{fig:RT_ctrl_8}
  \end{subfigure}\hfill
  \begin{subfigure}[b]{0.48\linewidth}
    \centering
    \includegraphics[width=\linewidth]{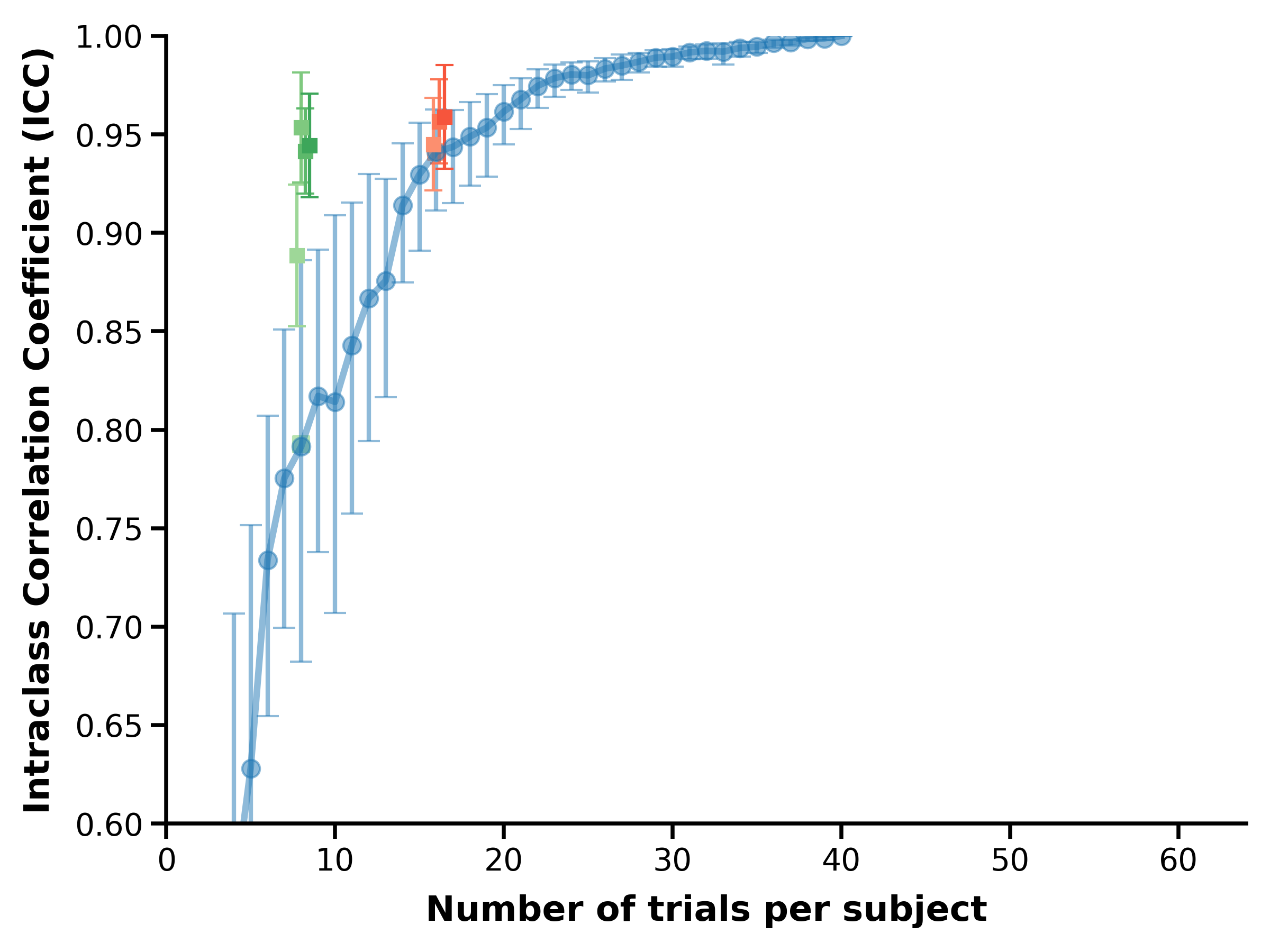}
    \caption{Control, 4-target protocol}
    \label{fig:RT_ctrl_4}
  \end{subfigure}

  \vspace{1ex}

  \begin{subfigure}[b]{0.48\linewidth}
    \centering
    \includegraphics[width=\linewidth]{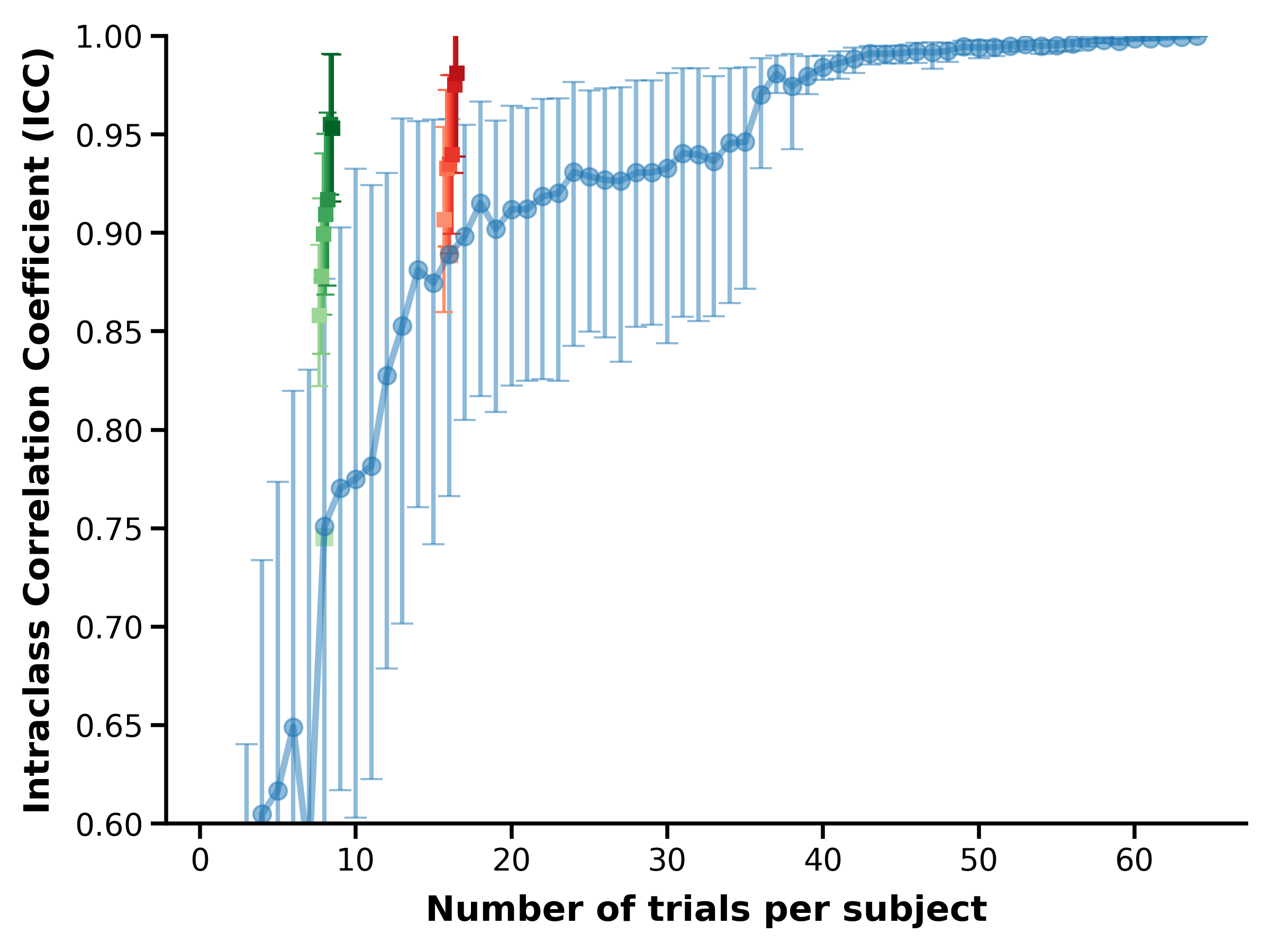}
    \caption{Stroke, 8-target protocol}
    \label{fig:RT_strk_8}
  \end{subfigure}\hfill
  \begin{subfigure}[b]{0.48\linewidth}
    \centering
    \includegraphics[width=\linewidth]{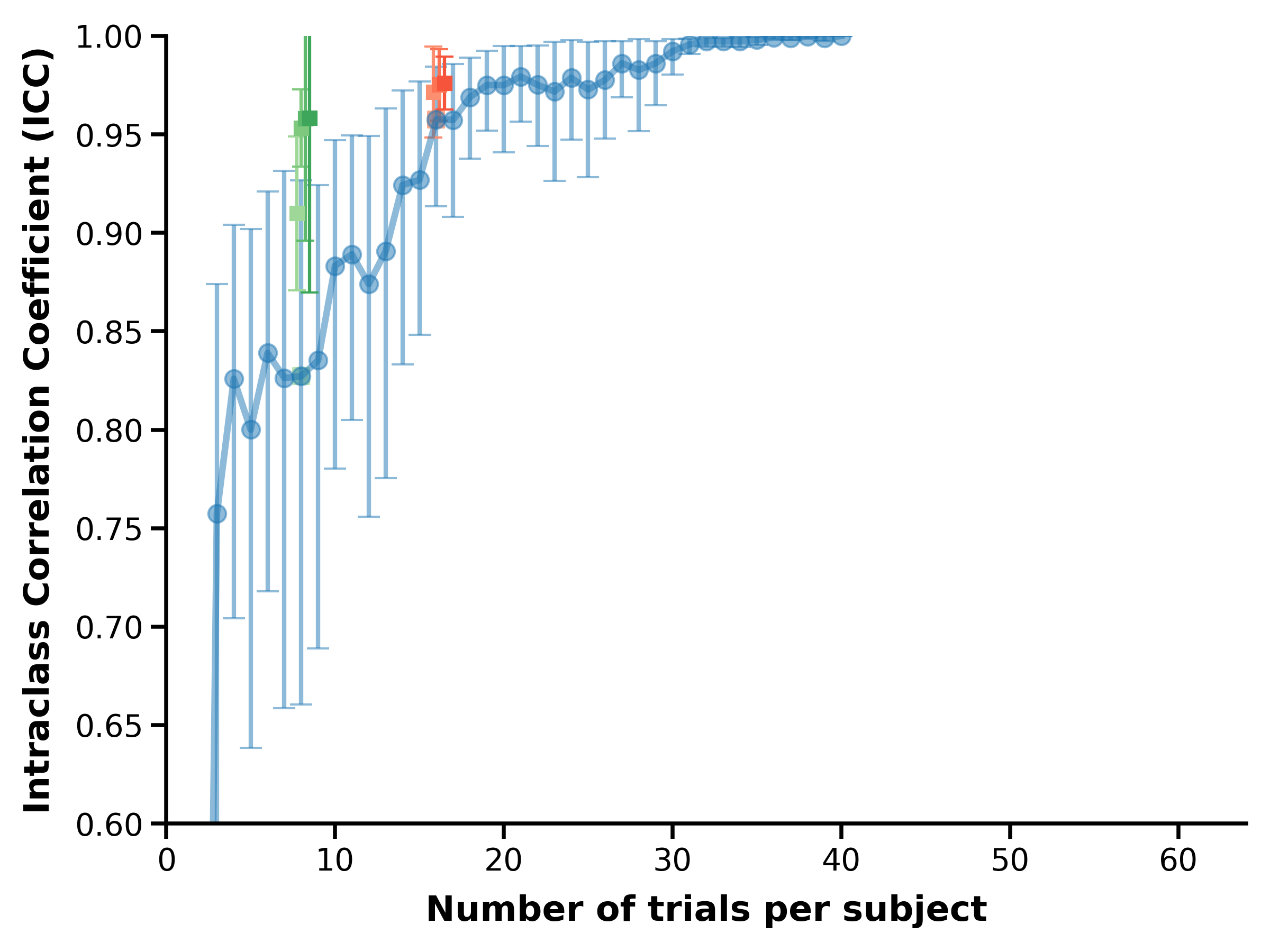}
    \caption{Stroke, 4-target protocol}
    \label{fig:RT_strk_4}
  \end{subfigure}

  \caption{Reaction time reliability (ICC) as a function of total trials for the Chronos model. (a) Control, 8-target; (b) Control, 4-target; (c) Stroke, 8-target; ; (d) Stroke, 4-target. Blue curve: ICC between  X-trial RT (median of X recorded trials) and complete-trial RT; error bars from subject bootstrap (B=1000). Squares: ICC for forecast-augmented protocols (8 recorded + k forecasts); error bars incorporate subject bootstrap and forecast-selection repeats. }

  \label{fig:RT_chronos_full}
\end{figure*}

\begin{figure*}[!htbp]
  \centering
  \begin{subfigure}[b]{0.48\linewidth}
    \centering
    \includegraphics[width=\linewidth]{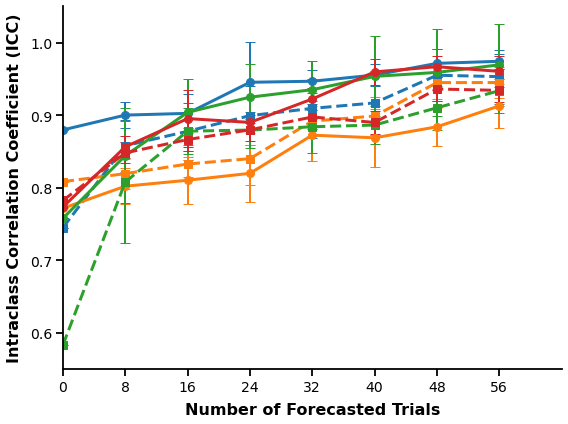}
    \caption{8-trial context, 8-target protocol}
    \label{fig:8trial-8target-ttc}
  \end{subfigure}\hfill
  \begin{subfigure}[b]{0.48\linewidth}
    \centering
    \includegraphics[width=\linewidth]{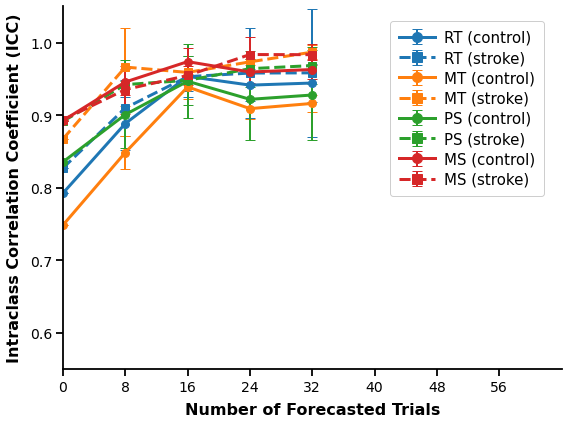}
    \caption{8-trial context, 4-target protocol}
    \label{fig:8trial-4target-ttc}
  \end{subfigure}

  \vspace{1ex}

  \begin{subfigure}[b]{0.48\linewidth}
    \centering
    \includegraphics[width=\linewidth]{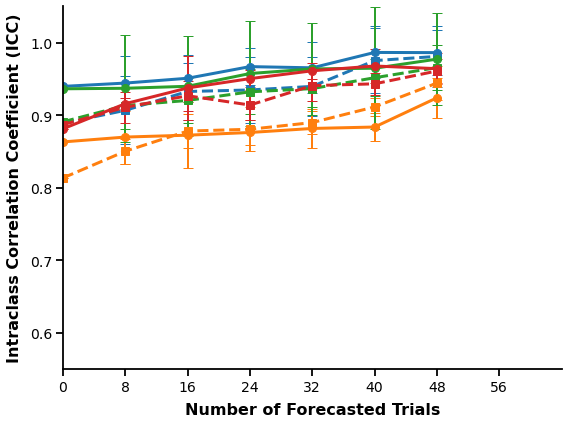}
    \caption{16-trial context, 8-target protocol}
    \label{fig:16trial-8target-ttc}
  \end{subfigure}\hfill
  \begin{subfigure}[b]{0.48\linewidth}
    \centering
    \includegraphics[width=\linewidth]{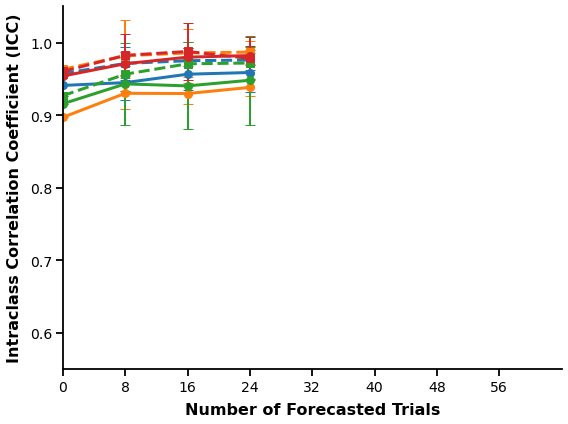}
    \caption{16-trial context, 4-target protocol}
    \label{fig:16trial-4target-ttc}
  \end{subfigure}

  \caption{Intraclass Correlation Coefficient (ICC) for reaction time (RT), movement time (MT), posture speed (PS), and max speed (MS), plotted as a function of the number of forecasted trials added to a fixed context. Each subplot corresponds to one of four task configurations: (a) 8-trial context, 8-target protocol; (b) 8-trial context, 4-target protocol; (c) 16-trial context, 8-target protocol; (d) 16-trial context, 4-target protocol. Solid lines with circular markers represent control participants; dashed lines with square markers represent participants with stroke. Colored lines indicate different kinematic parameters: blue = RT, orange = MT, green = PS, red = MS. The x-axis spans from 0 (context only) to 56 forecasted trials added in increments of 8, maintaining a consistent scale across all panels. Y-axis values denote ICC calculated from real + forecasted trials, with error bars representing standard deviation. The plots illustrate how forecasted trials affect reliability across groups, parameters, and task designs, with higher ICC generally achieved as more forecasted trials are included—though the extent of improvement varies by parameter and cohort.}
  \label{fig:target-trail-comp}
\end{figure*}

\begin{table}[!htbp]
  \centering
  \small
  \caption{Forecasting performance for the 8-target VGR protocol.}
  \label{tab:8target_grouped}
  \begin{tabular}{l l l ccc ccc}
    \toprule
    \textbf{Parameter} & \textbf{Cohort} & \textbf{Method}
      & \multicolumn{3}{c}{Context = 8} & \multicolumn{3}{c}{Context = 16} \\
    \cmidrule(r){4-6}\cmidrule(l){7-9}
    & & & ICC@8 & Best & $\Delta$ & ICC@16 & Best & $\Delta$ \\
    \midrule
    \multirow{6}{*}{Reaction Time}
      & \multirow{3}{*}{Control}
        & Chronos & 0.88 & 0.97 & 0.09 & 0.94 & 0.99 & 0.05 \\
      & & MOMENT & 0.88 & 0.93 & 0.05 & 0.94 & 0.96 & 0.02 \\
      & & ARIMA  & 0.88 & 0.90 & 0.02 & 0.94 & 0.93 & 0.01 \\
    \cmidrule{2-9}
      & \multirow{3}{*}{Stroke}
        & Chronos & 0.75 & 0.95 & 0.21 & 0.89 & 0.98 & 0.09 \\
      & & MOMENT & 0.75 & 0.90 & 0.15 & 0.89 & 0.95 & 0.06 \\
      & & ARIMA  & 0.75 & 0.85 & 0.10 & 0.89 & 0.90 & 0.01 \\
    \addlinespace
    \multirow{6}{*}{Movement Time}
      & \multirow{3}{*}{Control}
        & Chronos & 0.77 & 0.91 & 0.14 & 0.86 & 0.92 & 0.06 \\
      & & MOMENT & 0.77 & 0.89 & 0.12 & 0.86 & 0.90 & 0.04 \\
      & & ARIMA  & 0.77 & 0.85 & 0.08 & 0.86 & 0.88 & 0.02 \\
    \cmidrule{2-9}
      & \multirow{3}{*}{Stroke}
        & Chronos & 0.81 & 0.95 & 0.14 & 0.81 & 0.94 & 0.13 \\
      & & MOMENT & 0.81 & 0.90 & 0.09 & 0.81 & 0.92 & 0.11 \\
      & & ARIMA  & 0.81 & 0.88 & 0.07 & 0.81 & 0.90 & 0.09 \\
    \addlinespace
    \multirow{6}{*}{Posture Speed}
      & \multirow{3}{*}{Control}
        & Chronos & 0.76 & 0.97 & 0.21 & 0.94 & 0.98 & 0.04 \\
      & & MOMENT & 0.76 & 0.95 & 0.19 & 0.94 & 0.96 & 0.02 \\
      & & ARIMA  & 0.76 & 0.90 & 0.14 & 0.94 & 0.92 & -0.02 \\
    \cmidrule{2-9}
      & \multirow{3}{*}{Stroke}
        & Chronos & 0.58 & 0.93 & 0.35 & 0.89 & 0.97 & 0.07 \\
      & & MOMENT & 0.58 & 0.88 & 0.30 & 0.89 & 0.94 & 0.05 \\
      & & ARIMA  & 0.58 & 0.82 & 0.24 & 0.89 & 0.90 & 0.01 \\
    \addlinespace
    \multirow{6}{*}{Max Speed}
      & \multirow{3}{*}{Control}
        & Chronos & 0.77 & 0.96 & 0.19 & 0.88 & 0.96 & 0.08 \\
      & & MOMENT & 0.77 & 0.94 & 0.17 & 0.88 & 0.94 & 0.06 \\
      & & ARIMA  & 0.77 & 0.90 & 0.13 & 0.88 & 0.91 & 0.03 \\
    \cmidrule{2-9}
      & \multirow{3}{*}{Stroke}
        & Chronos & 0.78 & 0.93 & 0.15 & 0.89 & 0.96 & 0.07 \\
      & & MOMENT & 0.78 & 0.90 & 0.12 & 0.89 & 0.95 & 0.06 \\
      & & ARIMA  & 0.78 & 0.86 & 0.08 & 0.89 & 0.92 & 0.03 \\
    \bottomrule
  \end{tabular}
\end{table}

\begin{table}[!htbp]
  \centering
  \small
  \caption{Forecasting performance for the 4-target VGR protocol.}
  \label{tab:4target_grouped}
  \begin{tabular}{l l l ccc ccc}
    \toprule
    \textbf{Parameter} & \textbf{Cohort} & \textbf{Method}
      & \multicolumn{3}{c}{Context = 8} & \multicolumn{3}{c}{Context = 16} \\
    \cmidrule(r){4-6}\cmidrule(l){7-9}
    & & & ICC@8 & Best & $\Delta$ & ICC@16 & Best & $\Delta$ \\
    \midrule
    \multirow{6}{*}{Reaction Time}
      & \multirow{3}{*}{Control}
        & Chronos & 0.79 & 0.94 & 0.15 & 0.94 & 0.96 & 0.02 \\
      & & MOMENT & 0.79 & 0.90 & 0.11 & 0.94 & 0.94 & 0.00 \\
      & & ARIMA  & 0.79 & 0.88 & 0.09 & 0.94 & 0.92 & -0.02 \\
    \cmidrule{2-9}
      & \multirow{3}{*}{Stroke}
        & Chronos & 0.83 & 0.96 & 0.13 & 0.96 & 0.98 & 0.02 \\
      & & MOMENT & 0.83 & 0.92 & 0.09 & 0.96 & 0.96 & 0.00 \\
      & & ARIMA  & 0.83 & 0.88 & 0.05 & 0.96 & 0.92 & -0.04 \\
    \addlinespace
    \multirow{6}{*}{Movement Time}
      & \multirow{3}{*}{Control}
        & Chronos & 0.75 & 0.92 & 0.17 & 0.90 & 0.94 & 0.04 \\
      & & MOMENT & 0.75 & 0.88 & 0.13 & 0.90 & 0.92 & 0.02 \\
      & & ARIMA  & 0.75 & 0.85 & 0.10 & 0.90 & 0.90 & 0.00 \\
    \cmidrule{2-9}
      & \multirow{3}{*}{Stroke}
        & Chronos & 0.87 & 0.99 & 0.12 & 0.96 & 0.99 & 0.02 \\
      & & MOMENT & 0.87 & 0.95 & 0.08 & 0.96 & 0.98 & 0.02 \\
      & & ARIMA  & 0.87 & 0.90 & 0.03 & 0.96 & 0.94 & -0.02 \\
    \addlinespace
    \multirow{6}{*}{Posture Speed}
      & \multirow{3}{*}{Control}
        & Chronos & 0.84 & 0.93 & 0.09 & 0.92 & 0.95 & 0.03 \\
      & & MOMENT & 0.84 & 0.90 & 0.06 & 0.92 & 0.93 & 0.01 \\
      & & ARIMA  & 0.84 & 0.88 & 0.04 & 0.92 & 0.92 & 0.00 \\
    \cmidrule{2-9}
      & \multirow{3}{*}{Stroke}
        & Chronos & 0.89 & 0.96 & 0.07 & 0.93 & 0.97 & 0.04 \\
      & & MOMENT & 0.89 & 0.85 & 0.04 & 0.93 & 0.94 & 0.01 \\
      & & ARIMA  & 0.89 & 0.80 & -0.09 & 0.93 & 0.90 & -0.03 \\
    \addlinespace
    \multirow{6}{*}{Max Speed}
      & \multirow{3}{*}{Control}
        & Chronos & 0.89 & 0.96 & 0.07 & 0.95 & 0.98 & 0.03 \\
      & & MOMENT & 0.89 & 0.94 & 0.05 & 0.95 & 0.96 & 0.01 \\
      & & ARIMA  & 0.89 & 0.90 & 0.01 & 0.95 & 0.94 & -0.01 \\
    \cmidrule{2-9}
      & \multirow{3}{*}{Stroke}
        & Chronos & 0.89 & 0.98 & 0.09 & 0.96 & 0.98 & 0.02 \\
      & & MOMENT & 0.89 & 0.95 & 0.06 & 0.96 & 0.98 & 0.02 \\
      & & ARIMA  & 0.89 & 0.90 & 0.01 & 0.96 & 0.94 & -0.02 \\
    \bottomrule
  \end{tabular}
\end{table}

\subsection{Baseline reliability}

Figure \ref{fig:RT_chronos_full} illustrates the reliability of each Kinarm parameter (Reaction Time, Movement Time, Posture Speed, and Max Speed) as a function of the number of recorded trials, using ICCs calculated through bootstrapping with 95\% confidence intervals. To visualize the effect of forecast augmentation, square markers are overlaid on the same ICC–versus–trial-count curves: the lightest square denotes the ICC from the eight recorded ``context" trials (no augmentation), and progressively darker squares indicate the ICC achieved after adding forecast-generated trials in blocks of eight (i.e., +8, +16, +24, \ldots). Vertical error bars on these squares show the standard deviation across Monte Carlo–dropout passes used to generate forecasts. The horizontal axis for the 4-target protocol has been extended to align scale consistency across subplots, facilitating direct comparison of reliability metrics across task configurations.
For control participants, the ICC curves demonstrate that median ICC exceeded the excellent threshold ($\geq$ 0.90 per Koo and Li, 2016~\cite{KooLi2016}) after approximately 10–14 recorded trials across all parameters and task variants. For participants with stroke, reaching similar reliability required an additional 4–6 recorded trials. These findings establish baseline benchmarks against which forecast-augmented strategies are compared.

\subsection{Reliability with forecasts}
Figure \ref{fig:target-trail-comp} comprehensively displays ICC trajectories when combining real trials with Chronos forecasted data across participant groups, task variants, and kinematic parameters. Each subplot consistently ranges from 0 to 56 forecasted trials (increments of 8), providing an intuitive visualization of reliability changes as forecasts are progressively added.
Notably, Chronos forecasts substantially enhanced reliability across parameters and task variants compared to when only the actual 8 trials are sampled. For instance, panel (a) highlights that control participants, under the 8-target protocol with an initial 8-trial context with forecasted trials, achieved ICC values between 0.90–0.97 for all parameters with forecast augmentation—effectively matching reliability levels typically observed only after recording 24–28 actual trials. In participants with stroke under the same protocol (panel b), Reaction Time and Movement Time rapidly achieved ICC~$\geq$~0.90 with as few as eight forecasted trials appended to an 8-trial context. Posture Speed and Max Speed for participants with stroke also significantly benefited from forecasts, achieving comparable ICC values after a slightly expanded (16-trial) context.

\subsection{Method comparison}

Tables \ref{tab:8target_grouped} and \ref{tab:4target_grouped} summarize forecasting precision for Chronos, MOMENT, and ARIMA at the clinically relevant checkpoints of eight and sixteen recorded trials. For each method and context length, we report the following:
— ICC\@8/16, the reliability attainable when forecasts replace the remaining recorded trials;
— Best ICC, the maximum reliability observed anywhere along the forecast-augmented curve;
— $\Delta$ICC, the gain relative to the baseline ICC at the same context length.

Chronos delivered the largest $\Delta$ICC in 29 of the 32 parameter–cohort–protocol cells (91 \%), with typical gains of 0.10–0.21 for controls and 0.12–0.35 for stroke.  MOMENT closed roughly half that gap, whereas ARIMA rarely exceeded a 0.10 ICC improvement and sometimes degraded reliability (e.g.\ Posture Speed \emph{stroke}, 4-target: $\Delta$ICC = –0.02).

\section{Discussion}\label{sec:discussion}


This study is the first to demonstrate that a foundation-model approach can improve kinematic-based assessment of motor performance associated with stroke.  Chronos, fine-tuned on only 70 \% of our mixed stroke/control cohort and prompted with eight context trials, lifted ICC to within 5 points of the full-length reference for three of the four canonical parameters in controls and two in stroke (Figures \ref{fig:RT_chronos_full}, Table \ref{tab:8target_grouped}).  With a 16-trial context the gap narrowed to $\Delta\!{\rm ICC}\le 0.07$ for \emph{all} parameters and cohorts, effectively restoring assessment fidelity while discarding 60–75 \% of the original recording time.

MOMENT inherited the same pretraining recipe as Chronos but lagged by $\sim\!0.03$–0.06 ICC points in most scenarios, mirroring the relative ranking reported in the open-data Time-Bench benchmark \cite{Goswami2024MOMENT}.  The classical ARIMA baseline provided a useful lower bound: although its point forecasts were accurate for trajectories with highly stereotyped speed profiles, the model did not capture trial-to-trial variability and therefore under-estimated between-subject variance, yielding markedly smaller best-case ICCs (Tables \ref{tab:8target_grouped}, \ref{tab:4target_grouped}).  Together these results confirm that modern transformer-based forecasters—and not autoregressive statistics—are required to synthesise physiologically plausible reaches.


Limiting VGR to a compact recorded context and substituting model forecasts reduces the number of physically executed reaches while leaving the scoring pipeline unchanged. In practice, reducing the 8-target protocol from 64 recorded trials to eight recorded trials plus forecasts yields large session-time savings ($\geq$80\%; Fig.~\ref{fig:time-comp}). Crucially, the control–stroke separation in session-time distributions is preserved, indicating that forecast-based truncation does not bias group comparisons or reorder protocols. Together with the reliability results reported above, this supports the clinical viability of forecast-augmented assessment: it sharply cuts on-robot time without compromising interpretability or reliability.

Chronos’ superior performance likely stems from three architectural choices.
First, its tokenisation scheme discretises continuous signals into a ``time series vocabulary" that retains fine temporal detail while enabling powerful masked-language pretraining.
Second, the model’s relative multi-scale attention spans hundreds of samples, letting it capture both the bell-shaped speed pulse and the low-frequency curvature inherent in human reaches.
Third, the foundation-model objective yields a rich uncertainty estimate that, when sampled via Monte-Carlo dropout, produces a diverse pool of candidate trials.  Selecting subsets of that pool (green vs.\ orange markers in Figures \ref{fig:RT_chronos_full}) shows that diversity, not just mean accuracy, is critical for restoring ICC.

The concept of forecast-augmented assessment challenges the long-standing assumption that every trial must be physically executed. Once a model is fine-tuned to a dataset, subsequent patients could complete only the context block, after which the clinician would receive both standard KST metrics; in our evaluation we summarize uncertainty via error bars that reflect between-subject sampling and forecast-selection variability (See \ref{subsec:eval}). This pipeline could increase throughput in busy clinics, reduce fatigue in highly impaired populations who struggle with lengthy assessments, and enables adaptive testing: if early forecasts indicate low reliability, the robot can request a few extra real trials and stop as soon as the ICC target is met.

Prior efforts focused on shortening Kinarm assessments by recording fewer trials or skipping predictable tasks, rather than forecasting missing trajectories. For VGR specifically, Mostafavi \emph{et al.} showed that a reduced set of reaches can preserve SVM-based stroke/control discrimination from standard speed-derived parameters \cite{mostafavi-vgr-reduction}. At the battery level, hierarchical task ordering and cross-task similarity were used to infer outcomes and skip redundant tasks, yielding substantial time savings \cite{mostafavi2014hierarchical,mostafavi2015evaluation,Mostafavi2017}. In all cases, time is saved by selecting or reordering a fixed subset of physically recorded tasks/trials and discarding the rest. However, this approach simply considered whether a similar number of individuals were identified as impaired and not how trial reduction impacted assessment precision. 

In contrast, we \emph{forecast} the missing or noncompleted trials from as few as eight real reaches and then recompute the full suite of Kinarm parameters on the union of real + synthetic trials.  This enables us to preserve the diagnostic richness of the entire task (all eight movement directions, continuous speed profiles, endpoint stability, etc.) while reducing on-robot time.  Moreover, by generating multiple Monte Carlo–dropout realizations we explicitly model trial-level uncertainty, something that fixed-subset methods cannot capture. By sampling dropout-induced variations, we both reflect our confidence in each forecast and supply a richer ensemble of trajectories for more robust downstream feature estimation.

\subsection{Limitations}

Several caveats merit discussion. First, although we simulated prospective use by withholding 30 \% of participants for testing, genuine deployment will inevitably encounter domain shifts—new robot firmware, alternate seating configurations, or even different Kinarm installations, which may degrade out-of-the-box performance; lightweight, continual finetuning or federated updating strategies will likely be needed to maintain precision. Second, while our foundation models excel at reproducing aggregate kinematic statistics, they remain black boxes with respect to the underlying biomechanical strategies; integrating them with biomechanically constrained decoders or musculoskeletal simulators could yield more interpretable, in silico trials.

\subsection{Future directions}

Several avenues for extending this work merit exploration. First, the same forecasting framework could be applied to other Kinarm tasks, reverse Visually Guided Reaching \cite{lowrey2022impairments} or Ball on Bar \cite{lowrey2014novel}, so that the entire twelve-task KST battery might be completed in less time, improving clinical throughput. Second, because our Monte Carlo–dropout approach produces not a single ``best guess" trajectory but a distribution of plausible future reaches, it becomes possible to visualise each patient’s individualized performance envelope. Clinicians could use these envelopes to set personalised difficulty thresholds for rehabilitation programs, selecting targets or perturbations that sit just beyond a patient’s lower confidence bound, thereby delivering rehabilitation exercises optimally tuned to each individual’s capacity and uncertainty profile. Finally, while our finetuning leveraged a few hundred participants, foundation models promise even broader generality when pretrained on massive movement corpora. By ingesting millions of hours of daily-life accelerometer or wearable-sensor recordings, one could endow these models with priors that capture the full spectrum of human motion. Such large-scale pretraining would not only improve forecasts for stroke rehabilitation but also likely transfer to other movement disorders such as Parkinson’s disease, cerebral palsy, or multiple sclerosis.

\section{Conclusion}

This study targeted the assessment time burden of full‑length Kinarm VGR blocks (40–64 trials), which can be fatiguing for some stroke survivors. We investigated whether time‑series foundation models (Chronos, MOMENT) could forecast the missing trials from only 8–16 recorded reaches while preserving the reliability of VGR parameters. Using a forecast‑augmented protocol, we showed that Chronos, fine‑tuned on our mixed control–stroke cohort, recovers ICCs to within 0.05–0.07 of the full‑length reference (parameter/cohort dependent) while reducing session time by 75–88\% for the most impaired participants with stroke; MOMENT underperformed Chronos, and ARIMA provided only modest gains. To our knowledge, this is the first demonstration that foundation models can safely replace a large fraction of physically recorded robotic reaches in a reaching task.

\backmatter

\bmhead{Acknowledgements}

We would like to sincerely thank Simone Appaqaq, Kimberly Moore, Ethan Heming, Mary-Jo Demers, Helen Bretzke, and Justin Peterson for their valuable assistance with participant recruitment and assessment, database management, and technical support.

\section*{Declarations}

\begin{itemize}
\item \textbf{Funding}
This research was funded by a Canadian Institutes of Health Research Operating Grant (MOP 106662), a Heart and Stroke Foundation of Canada Grant-in-Aid (G-13-0003029), an Alberta Innovates Health Solutions Team Grant (201500788), an Ontario Research Fund – Research Excellence grants (ORF-RE 04-47, RE-09-112), Vector AI Institute, a Canada CIFAR AI Chair, Natural Sciences and Engineering Research Council of Canada, and Queen’s University.
\item \textbf{Conflict of interest/Competing interests} SHS is co-founder and CSO of Kinarm (formally known
as BKIN Technologies), the company that commercializes the robotic technology used in the present study. All other authors confirm no conflict of interest.
\item \textbf{Ethics approval and consent to participate} This study was reviewed and approved by the Queen’s University Research Ethics Board and the University of Calgary Conjoint Health Research Ethics Board. All participants gave their written informed consent to have their data collected for research purposes before performing the assessment.
\item \textbf{Consent for publication} Consent for publication was obtained from all participants.
\item \textbf{Data availability} The dataset generated for the present study is not accessible to the public. However, data can be obtained from SHS (steve.scott@queensu.ca) upon reasonable request.
\item \textbf{Materials availability} Not applicable
\item \textbf{Code availability} All Python code used in the analysis of this data is available on https://github.com/FaranakAk
\item \textbf{Author contribution} \noindent FA conceived the concept and methodology, analyzed data, contributed to the interpretation of results, drafted the manuscript, and participated in the editing process. 

\noindent NM contributed to methodology, interpretation of results, and partcipated in editing the manuscript.

\noindent SPD coordinated the collection of data, contributed to task design, participated in the editing process.

\noindent SHS coordinated the collection of data, contributed to task design, provided the concept and methodology, contributed to the interpretation of results, participated in the editing process.

\noindent PM conceived the concept and methodology, contributed to the interpretation of results, participated in the editing process.

\noindent All authors reviewed and approved the final manuscript before its submission.
\end{itemize}

\noindent






\begin{appendices}






\end{appendices}


\bibliography{sn-bibliography}

\end{document}